\documentclass[11pt]{article}

\usepackage{times}
\usepackage{latexsym}
\usepackage{amsmath}
\usepackage{fullpage}
\usepackage[T1]{fontenc}
\usepackage[utf8]{inputenc}
\usepackage{microtype}

\usepackage{natbib}
\usepackage{hyperref}
\usepackage{url}
\usepackage{graphicx}
\usepackage{xcolor}
\newcommand{\arxiv}[1]{#1}
\newcommand{\conf}[1]{}

\title{Scaling Language Model Size in Cross-Device Federated Learning}
\author{Jae Hun Ro$^*$ \and Theresa Breiner \and Lara McConnaughey 
\and Mingqing Chen \and Ananda Theertha Suresh \and Shankar Kumar \and Rajiv Mathews}
 \date{ Google \\[2ex]
\texttt{$^*$jaero@google.com} 
}

\begin{document}
\maketitle

\begin{abstract} 
Most studies in cross-device federated learning focus on small models, due to the server-client communication and on-device computation bottlenecks. In this work, we leverage various techniques for mitigating these bottlenecks to train larger language models in cross-device federated learning. With systematic applications of partial model training, quantization, efficient transfer learning, and communication-efficient optimizers, we are able to train a $21$M parameter Transformer and $20.2$M parameter Conformer that achieve the same or better perplexity as that of a similarly sized LSTM with $\sim10\times$ smaller client-to-server communication cost and $11\%$ lower perplexity than smaller LSTMs commonly studied in literature.
\end{abstract}

\section{Introduction}

Federated learning is a distributed training technique, where a model is trained on data distributed across clients or edge devices without user-generated data ever leaving the device, providing an additional layer of privacy and security \citep{konevcny2016federated,konecny2016federated2, mcmahan2017communication}.
We refer readers to \cite{li2019federated, kairouz2019advances} for a detailed literature survey on federated learning.
Federated learning has been used in several applications including virtual keyboard applications \citep{hard2018federated}, keyword spotting \citep{fedkeyword2020}, and healthcare \citep{brisimi2018federated}.

Language models (LM) have many uses in language-based applications including virtual keyboard \citep{chen-etal-2019-federated, Zhang2021PositionInvariantTW} and automatic speech recognition 
\citep{kannan2018externallm,variani2020hybrid,conformerlm}.
Recently, there has been increased interest in training progressively larger and deeper LMs with impressive quality improvements in downstream tasks, including question answering, text classification, and text summarization \citep{devlin-etal-2019-bert,dai-etal-2019-transformer,zhilin2019xlnet,irie2019deeplmtransformer,kaplan2020scaling}.
These models tend to be variants of the Transformer \citep{vaswani2017}. Recently, Conformer models, which employ convolution layers in Transformer-based architectures, have also been proposed \citep{gulati20_interspeech}.

Federated learning is typically studied in two scenarios: \emph{cross-silo}, where the number of clients is small, and \emph{cross-device}, where the number of clients can be in the order of millions \citep{hard2018federated}.
In this work we focus on cross-device, where devices are typically edge devices such as cell phones, with limited computation and communication capabilities.
Hence, the major benchmark LMs tend to be very limited in size \citep{mcmahan2017communication,mcmahan2018learning, caldas2019leaf, reddi2020adaptive,sim21_interspeech} because memory, computation, and communication are critical bottlenecks \citep{kairouz2019advances}.
In particular, previous works that train federated LMs in production settings have used coupled input forget gate (CIFG) long short-term memory (LSTM) models with fewer than 4 million parameters \citep{hard2018federated,chen-etal-2019-federated,ramaswamy2020training}. 
These resource constraints have motivated research into various efficient algorithms for training larger models with federated learning \citep{konevcny2016federated,hamer2020fedboost}.
However, most of these techniques are still evaluated on relatively small models compared to their server-based counterparts.
In this work, we systematically evaluate multiple strategies for mitigating communication and computation costs of training larger LMs to determine if the impressive quality gains from larger models can also be achieved in cross-device federated learning. 

While there are previous works on \emph{efficient} Transformers \citep{tay2020efficient,tay2021long}, we forgo these efficient variants as they may actually be more inefficient when sequences are short \citep{katharopoulos2020transformers,choromanski2021rethinking}.
Additionally, \citet{lin2020ensemble, liu2020federated, hilmkil2021scaling} trained large Transformer models in the cross-silo setting, where devices have more resources, whereas we focus on the resource-constrained cross-device setting. 

Recent large LMs, such as GPT-3 \cite{gpt3}, contain hundreds of billions of parameters, which is substantially bigger than the memory limits of edge devices.
Therefore in this work, we consider \emph{large} models to be at most $25$ million parameters, which is still considerably larger than existing models trained on-device. 

The rest of the paper is organized as follows. In Section~\ref{sec:contrib}, we overview our contributions.
In Section~\ref{sec:data_model}, we detail the dataset and models.
We then analyze techniques to reduce the per-round cost in Section~\ref{sec:per_round_cost}, and the number of communication rounds in Section~\ref{sec:num_rounds}.
Finally in Section~\ref{sec:combination}, we combine techniques and demonstrate that large Transformers can be trained using many fewer rounds and significantly lower communication and computation cost. 

\section{Our contributions}
\label{sec:contrib}

We explore two regimes: small models typically studied in cross-device federated learning with fewer than $5$M parameters and new larger models with at most $25$M parameters. We study three architectures: CIFG-LSTM \citep{hochreiter1997}, or LSTM for simplicity, \citep{hard2018federated}, Transformer \citep{vaswani2017}, and Conformer \citep{gulati20_interspeech}. We refer to both the Transformer and Conformer as Transformer-based models. Our contributions are the following:

\begin{itemize}
    \item We are the first to investigate Transformer-based LMs with 25M parameters for cross-device federated learning, which we find outperform LSTMs of similar size.
    \item We demonstrate that large models substantially outperform small models on standard tasks but at much higher communication and computation costs, requiring $4\times$ the communication cost per round.
    \item We investigate quantization and partial model training to address the per round communication and computation cost. With quantization, we achieve similar perplexity with half the download cost and one quarter of the upload cost, reducing total communication cost by $62.5\%$. Partial model training can further reduce the upload cost by $70\%$.
    \item We study transfer learning as a method of reducing the number of communication rounds and show that centralized pretraining on a suitable alternate corpus reduces the total communication rounds by $3\times$.
    \item We show that the combination of above techniques can be used to train a Large Transformer and Conformer with the same perplexity as that of a similarly sized LSTM with $\sim 10\times$ the smaller client-to-server communication cost. 
\end{itemize}

\section{Dataset and models}
\label{sec:data_model}
In this section, we describe the models and dataset used in the rest of the paper.
We train on the Stack Overflow federated dataset from \citet{tff}, which contains posts from the public forum grouped by username. 
Following trends in training Transformers, we use sentence-piece \citep{kudo-richardson-2018-sentencepiece} for sub-word tokenization with a vocabulary size of $4$K.
The sentence-piece model is computed based on the entire Stack Overflow training corpus in an offline process on server.
During federated learning, this fixed sentence-piece model is transmitted to each client to encode the local text data.
Doing so provides greater coverage for cross-dataset applications as well as potential downstream speech applications such as ASR \cite{li2021,sim21_interspeech}.
We measure performance on next-subword prediction using test perplexity.
See Appendix~\ref{app:data_model} for descriptive dataset statistics. 
All experiments were implemented using JAX \citep{jax2018github} and FedJAX \citep{ro2021fedjax} federated simulation libraries.

We first did a hyperparameter search for each model and size ($\leq5$M and $\leq25$M), with FedAdam \citep{reddi2020adaptive}, or FedAvg for simplicity, 
with $200$ clients per round for $3$K rounds, resulting in six models: \emph{Small LSTM} ($4.7$M), \emph{Large LSTM} ($18.8$M), \emph{Small Transformer} ($4.1$M), \emph{Large Transformer} ($21$M), \emph{Small Conformer} ($4.1$M), and \emph{Large Conformer} ($20.2$M).

\begin{figure}[h]
\centering
\arxiv{\includegraphics[scale=0.42]{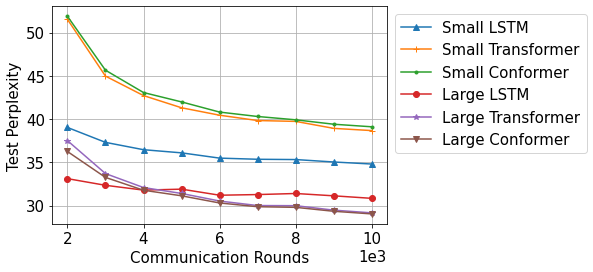}}
\conf{\includegraphics[scale=0.32]{so_fedavg.png}}
\caption{Test perplexity over communication rounds for each class and size of model.}
\label{fig:fedavg-baseline}
\end{figure}

We then trained the chosen architectures with $800$ clients per round for $10$K rounds in Figure~\ref{fig:fedavg-baseline}.
As expected, the larger variants significantly outperform their smaller counterparts with the Large Conformer achieving the best perplexity.
However, the larger models are more expensive to train per round and although the Large Conformer achieves the best perplexity, it only surpasses the Large LSTM after $4$K rounds.
Next, we focus on techniques to reduce this cost per round and number of rounds.
For more details about the architecture search, the selected models, and their performance, see Appendix~\ref{app:data_model}. 

\section{Cost per round}
\label{sec:per_round_cost}

The larger models have $18.8$M, $21$M, and $20.2$M parameters ($150$MB, $168$MB, and $162$MB at $32$ bits per parameter) which need to be downloaded, trained, and uploaded at each round, a strain on both communication and computation on device. There are often strict time or transfer byte limits for each round of training, which can prohibit some devices from training these models due to slower transfer/processing speeds \citep{kairouz2019advances}.
We show that we can significantly reduce these costs by partial model training and quantization techniques.

\textbf{Partial model training}: 
Training only a subset of the model can reduce the computational cost of training and has been examined in both federated \citep{caldas2019expanding,yang2021partial} and non-federated \citep{kovaleva-etal-2019-revealing} settings.
Additionally, reducing the number of trainable parameters can also decrease communication cost since only the trainable parameters need to be uploaded.

\begin{figure}[h]
\centering
\arxiv{\includegraphics[scale=0.42]{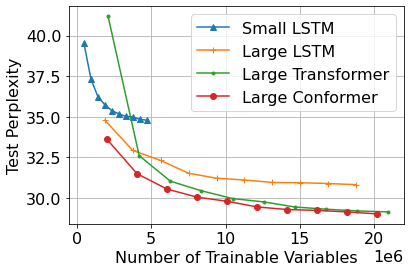}}
\conf{\includegraphics[scale=0.32]{so_pvt_trainable.png}}
\caption{Test perplexity as a function of number of trainable variables.}
\label{fig:pvt}
\end{figure}

We follow the Partial Variable Training (PVT) per client per round  strategy \citep{yang2021partial} as it only freezes a subset of the original model and can be applied generally to multiple model architecture types. For more experiment details, see Appendix~\ref{app:pvt}.
We report test perplexity as a function of number of trainable variables in Figure~\ref{fig:pvt}.
Large LSTM and Conformer seem to be able to handle more aggressive parameter freezing compared to Large Transformer in terms of quality regression.
Additionally, training only $30\%$ of variables for the Large Conformer ($6.1$M) achieves better performance than the full Large LSTM ($18.8$M).

\textbf{Quantization}: 
To reduce communication costs, various quantization strategies can decrease the number of bits required to represent model parameters \citep{bernstein2018signsgd,pmlr-v108-reisizadeh20a,gandikota2021vqsgd,vargaftik2021drive}. We examine stochastic k-level uniform quantization \citep{alistarh2017qsgd, suresh2017distributed} as it can be applied to model parameters on download (server-to-client) and model updates on upload (client-to-server) communication with adjustable levels of compression, and compare with TernGrad, an upload technique \citep{wen2017terngrad}. 

\begin{figure}[h]
\centering
\arxiv{
\includegraphics[scale=0.36]{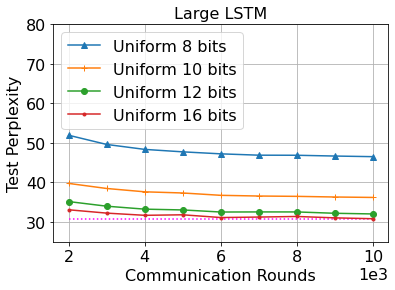}
\includegraphics[scale=0.36]{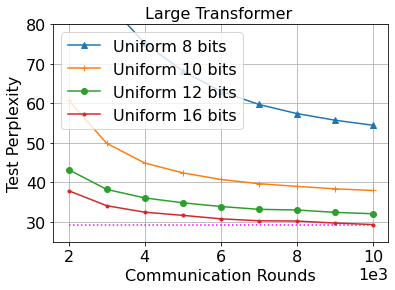}
\includegraphics[scale=0.36]{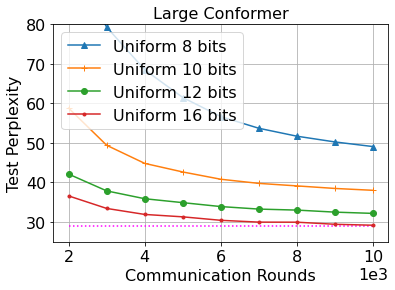}
}
\conf{
\includegraphics[scale=0.26]{rnn_quant_download.png}
\includegraphics[scale=0.26]{large_trans_download_quant_left_leg.png}
\includegraphics[scale=0.26]{conf_quant_download.png}
}
\caption{Test perplexity over communication rounds for varying download quantization levels, with upload quantization fixed to $8$ bits. Dashed line shows the baseline without quantization.}
\label{fig:quant_download}
\end{figure}

We focus analysis on larger models which are more affected by quantization. The LSTM appears more "quantizable" during download than the Transformer and Conformer, with less regression in Figure~\ref{fig:quant_download}. 
The perplexities of the Transformer and Conformer with $16$ download bits match that of their corresponding baselines and with $12$ bits are close to that of the LSTM.

\begin{figure}[h]
\centering
\arxiv{
\includegraphics[scale=0.36]{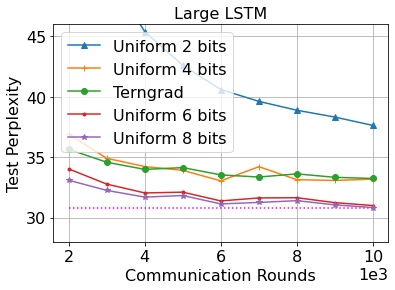}
\includegraphics[scale=0.36]{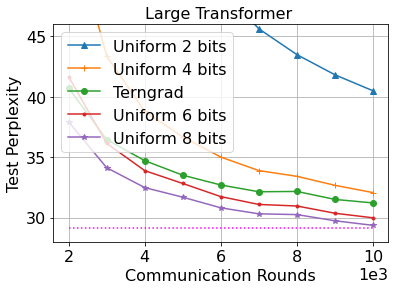}
\includegraphics[scale=0.36]{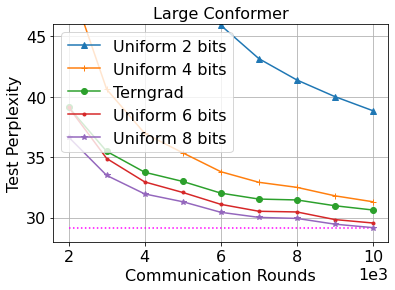}
}
\conf{
\includegraphics[scale=0.26]{rnn_upload_quant.png}
\includegraphics[scale=0.26]{trans_quant_upload.png}
\includegraphics[scale=0.26]{conf_quant_upload.png}
}
\caption{Test perplexity over communication rounds for varying upload quantization levels, with download quantization fixed to $16$ bits. TernGrad is comparable to uniform with about $1.6$ bits. Dashed line shows the baseline without quantization.}
\label{fig:quant_upload}
\end{figure}

\begin{figure}[t]
\centering
\arxiv{\includegraphics[scale=0.42]{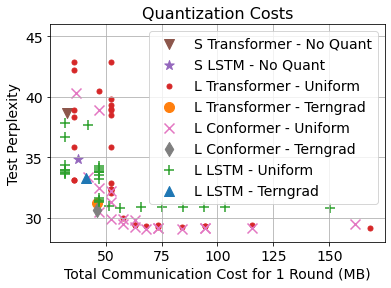}}
\conf{\includegraphics[scale=0.32]{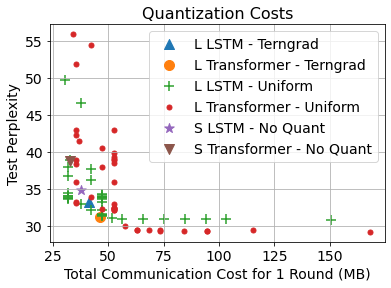}}
\caption{Test set perplexity versus total communication cost (download $+$ upload) in a single round of training, for each quantization algorithm. Uniform settings include points for varying quantization bits.}
\label{fig:quant_comm_costs}
\end{figure}

For all models, $8$ bit upload matches the corresponding baselines, or even $6$ bits for the LSTM in Figure~\ref{fig:quant_upload}. TernGrad, requiring $\log_2(3)$ bits, outperforms the $4$ bit in the Transformer and Conformer but not for the LSTM. It provides the best cost-performance tradeoff in Figure~\ref{fig:quant_comm_costs}.
More details are in Appendix~\ref{app:quant}.

\section{Number of communication rounds}
\label{sec:num_rounds}

\textbf{Transfer learning}: Transfer learning leverages pretrained models to improve model quality \citep{pmlr-v97-houlsby19a}.
By pretraining, the number of communication rounds required for model convergence can be significantly reduced \citep{stremmel2020pretrain}.

\begin{figure}[h]
\centering
\arxiv{
\includegraphics[scale=0.36]{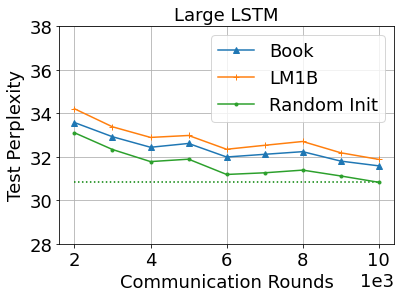}
\includegraphics[scale=0.36]{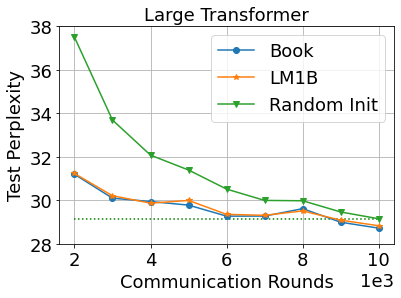}
\includegraphics[scale=0.36]{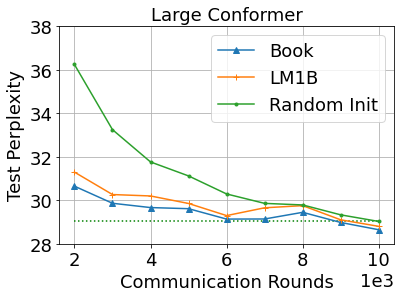}
}
\conf{
\includegraphics[scale=0.26]{large_lstm_pretrain.png}
\includegraphics[scale=0.26]{large_trans_pretrain.png}
\includegraphics[scale=0.26]{large_conf_pretrain.png}
}
\caption{Test perplexity over communication rounds comparing pretraining corpora. Dashed line is the final perplexity reached by the randomly initialized model.}
\label{fig:pretraining}
\end{figure}

We use two datasets for pretraining: a large corpus of digitized books \citep{Zhang2021PositionInvariantTW} and the One Billion Word Benchmark (LM1B) \citep{Chelba2014OneBW}.
After pretraining using synchronous SGD for $30$M steps, we finetune on Stack Overflow using FedAvg.
For additional details, see Appendix~\ref{app:transfer}.
We report results for each of the pretraining datasets and random initialization in Figure~\ref{fig:pretraining}.

Books consistently outperforms LM1B for all models.
Pretraining greatly benefits the Large Transformer and Conformer compared to the Large LSTM, reducing the number of rounds needed to reach the final $10$K without pretraining by $4$K rounds.
Furthermore, at round $2$K, the Large Transformer and Conformer already outperform the Large LSTM, making the number of rounds needed for training similar to that of smaller models used in mobile keyboard prediction \citep{hard2018federated}. 

\begin{figure}[h]
\centering
\arxiv{
\includegraphics[scale=0.36]{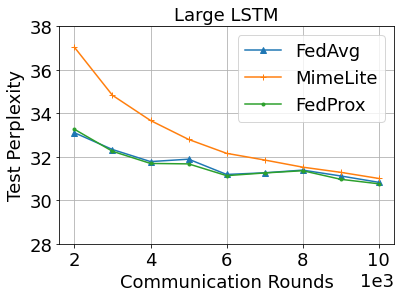}
\includegraphics[scale=0.36]{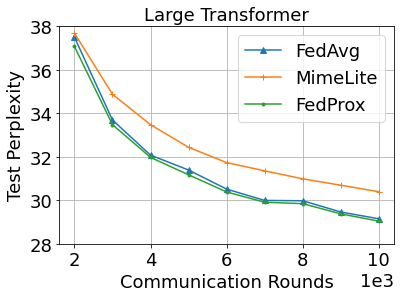}
\includegraphics[scale=0.36]{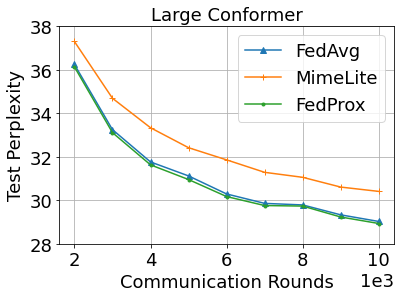}
}
\conf{
\includegraphics[scale=0.26]{so_lstm_opt.png}
\includegraphics[scale=0.26]{so_trans_opt.png}
\includegraphics[scale=0.26]{so_conf_opt.png}
}
\caption{Test perplexity over communication rounds for each model and algorithm.}
\label{fig:comm-opt}
\end{figure}

\textbf{Different optimizers}: 
Since the introduction of FedAvg, several variations continue to be developed \citep{li2018federated,hamer2020fedboost,reddi2020adaptive}.
Specifically, we examine MimeLite \citep{karimireddy2020mime} and FedProx \citep{li2018federated} as they have been shown to reduce the total amount of rounds required for provable convergence.
However, in Figure~\ref{fig:comm-opt}, FedProx and MimeLite do not improve convergence speed over FedAvg.
More details can be found in Appendix~\ref{app:comm-opt}.

\begin{figure}[t]
\centering
\arxiv{\includegraphics[scale=0.42]{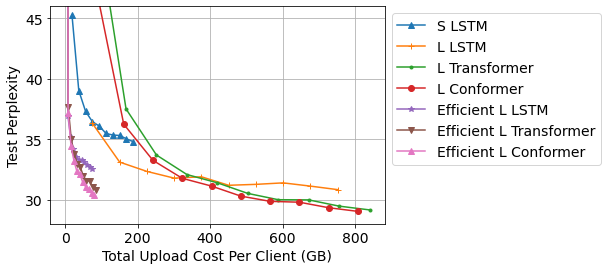}}
\conf{\includegraphics[scale=0.32]{so_combo.png}}
\caption{Test perplexity over total uploaded gigabytes per client for each class of model.}
\label{fig:combo-upload}
\end{figure}

\section{Combination of techniques} 
\label{sec:combination}

We experiment with combining partial model training, quantization, and transfer learning to train \emph{efficient} larger models.
For these experiments, we train on just $40\%$ of trainable parameters with PVT
and warm start after pretraining on the Books corpus.
Combining download quantization with these techniques did not perform as well, so we only apply $8$ bit uniform quantization on upload, which is the tightest communication bottleneck (\citet{mobile-speeds-05-2021} reports that mobile upload speeds worldwide are over $4\times$ slower than download as of May 2021).
For the full experiment details, refer to Appendix~\ref{app:combo}.
We report the test perplexity in terms of total upload communication cost in Figure~\ref{fig:combo-upload}.
Restricting for small upload costs ($<200$GB), the efficient models outperform all others with the efficient Large Conformer yielding the best perplexity.
Furthermore, the efficient Large Transformer and efficient Large Conformer achieve the same or better perplexity as the Large LSTM with no efficient techniques.

\section{Conclusion}

We systematically studied several techniques for addressing the communication and computation bottlenecks of federated learning.
We further demonstrated that these techniques, individually or in combination, can scale to larger models in cross-device federated learning.
Extending this study to other architectures and efficient strategies remains an interesting open question.

\newpage
\bibliographystyle{abbrvnat}
\bibliography{references}

\newpage
\appendix
\onecolumn
\begin{center}
    {\Large{Appendix}}
\end{center}

\section{Dataset and models}
\label{app:data_model}

\begin{figure}[h]
\centering
\includegraphics[scale=0.45]{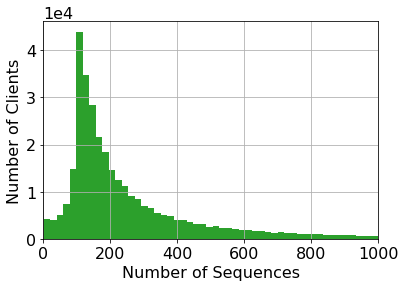}
\includegraphics[scale=0.45]{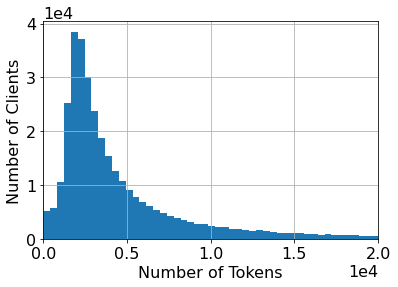}
\includegraphics[scale=0.45]{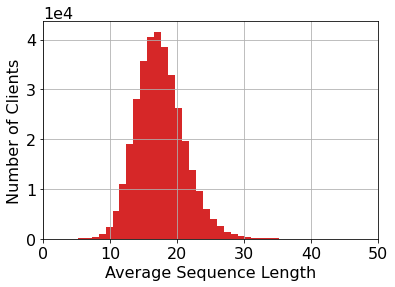}
\caption{Stack Overflow train split sub-word statistics.}
\label{fig:stackoverflow-stats}
\end{figure}

\begin{table}[h]
\centering
\caption{Selected architectures for each model and size range. The values in $[\ ]$ are the possible hyperparameter values searched over.
Layer Size refers to the LSTM layer dimension and MLP layer dimension for Transformer and \# Layers refers to number of LSTM layers and number of Transformer and Conformer blocks. Note that for the Conformer, the Layer Size is directly tied to the Embedding Size.}
\begin{tabular}{ccccc}
Model & \# Parameters & Embedding Size & Layer Size & \# Layers \\
& & $[128, 256, 512, 1024]$ & $[512, 1024, 2048]$ & $[1, 2, 3, 4, 6, 8]$ \\
\hline
Small LSTM & $4.7$M & $256$ & $2048$ & $1$ \\
Small Transformer & $4.1$M & $128$ & $2048$ & $6$ \\
Small Conformer & $4.1$M & 256 & $-$ & $2$ \\
\hline
Large LSTM & $18.8$M & $1024$ & $2048$ & $1$ \\
Large Transformer & $21.0$M & $512$ & $2048$ & $6$ \\
Large Conformer & $20.2$M & $512$ & $-$ & $3$ \\
\end{tabular}

\label{tab:arch-sweep}
\end{table}

\begin{table}[h]
\centering
\caption{Test metrics after $10$K rounds of training for each class of model and number of clients per round. The results in \textbf{bold} indicate the best for each size range.}
\begin{tabular}{ccc}
Model & \# Clients & Perplexity \\
\hline
Small LSTM & $200$ & $35.31$ \\
Small LSTM & $400$ & $34.93$ \\
Small LSTM & $800$ & $\mathbf{34.80}$ \\
\hline
Small Transformer & $200$ & $40.18$ \\
Small Transformer & $400$ & $39.38$ \\
Small Transformer & $800$ & $38.66$ \\
\hline
Small Conformer & $200$ & $38.22$ \\
Small Conformer & $400$ & $37.53$ \\
Small Conformer & $800$ & $36.80$ \\
\hline
\hline
Large LSTM & $200$ & $30.97$ \\
Large LSTM & $400$ & $30.79$ \\
Large LSTM & $800$ & $30.83$ \\
\hline
Large Transformer & $200$ & $30.64$ \\
Large Transformer & $400$ & $29.81$ \\
Large Transformer & $800$ & $29.15$ \\
\hline
Large Conformer & $200$ & $30.44$ \\
Large Conformer & $400$ & $29.66$ \\
Large Conformer & $800$ & $\mathbf{29.06}$ \\
\end{tabular}
\label{tab:baseline}
\end{table}

\begin{table}[h]
\centering
\caption{Selected hyperparameters for each model and size range.
The values in $[\ ]$ are the possible hyperparameter values searched over.
Batch Size, \# Examples, and Clipnorm here apply to the client local SGD steps. LR is learning rate.}
\begin{tabular}{cccccc}
Model & Batch Size & \# Examples & Clipnorm & Client LR & Server LR \\
& $[8, 16]$ & $[1200, 1600]$ & $[0.0, 16.0]$ & $[0.01, 0.1, 0.5, 1.0, 2.0]$ & $[0.001, 0.01]$ \\
\hline
Small LSTM & $16$ & $1200$ & $16.0$ & $1.0$ & $0.001$ \\
Small Transformer & $16$ & $1200$ & $0.0$ & $0.1$ & $0.001$ \\
Small Conformer & $16$ & $1200$ & $0.0$ & $0.1$ & $0.001$ \\
\hline
Large LSTM & $16$ & $1200$ & $16.0$ & $1.0$ & $0.001$ \\
Large Transformer & $16$ & $1200$ & $0.0$ & $0.5$ & $0.001$ \\
Large Conformer & $16$ & $1200$ & $0.0$ & $1.0$ & $0.001$ \\
\end{tabular}

\label{tab:baseline-hyper}
\end{table}

\begin{figure}
\centering
\includegraphics[scale=0.45]{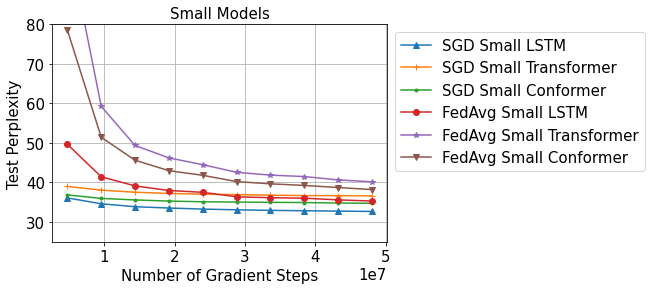}
\includegraphics[scale=0.45]{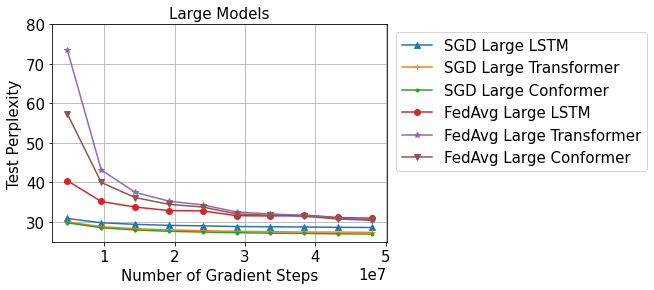}
\caption{Test set perplexity as a function of number of gradient computations for comparing the centralized and federated averaging baselines.}
\label{fig:fedavg-central-baseline}
\end{figure}

For the baseline architecture search, Table~\ref{tab:arch-sweep} details the selected architectures as well as the search ranges for each dimension.
The final hyperparameters were selected based on the test perplexity after $3$K rounds of training using FedAvg with $200$ clients per round.
From here on, we fix the Adam optimizer with $\beta_1$ at $0.9$, $\beta_2$ at $0.999$, and epsilon at $1e^{-8}$.
Additionally, based on the distribution of average sequence lengths across Stack Overflow clients in Figure~\ref{fig:stackoverflow-stats}, we fix the max sequence length for training and evaluation to $30$.

Table~\ref{tab:baseline} contains the results for each selected model after $10$K rounds of training using FedAvg with $200$, $400$, and $800$ clients per round.
As expected, the best results are achieved by using $800$ clients per round.
Thus, from here on, we report results for $800$ clients per round only.
For these experiments, we also search over client learning rate, client batch size, client max number of examples (with client number of epochs fixed to $1$), client $\ell_2$ norm for clipping, and server learning rate.
The search ranges as well as selected values for each model are detailed in Table~\ref{tab:baseline-hyper}.
For all following experiments, we fix client batch size to $16$ and client max number of examples to $1200$ since the larger batch size consistently performed the best and Figure~\ref{fig:stackoverflow-stats} shows that $1200$ sequences is more than enough to cover the vast majority of clients with the number of epochs fixed at $1$.
We also search over the same ranges for all following experiments where applicable for consistency.

As an additional baseline comparison, we also train each model using synchronous SGD to observe model quality in terms of number of gradient computations.
These centralized baselines provide a rough estimate of an upper bound on model quality for federated learning.
To produce a reasonable comparison between the federated and centralized experiments, we compare by number of gradient computations.
We approximate the number of gradient steps taken for federated learning with $200$ clients per round for $10$K communication rounds.
We train the centralized models using the Adam optimizer and run periodic evaluation on the test set at the same frequency as the federated experiments.
We compare final metrics between centralized and federated training on the test set in Figure~\ref{fig:fedavg-central-baseline}.
Observing the test perplexity over gradient steps, it is evident that the relative rankings of the models remain consistent between centralized and federated baselines.
Additionally, by $10$K rounds, the large federated models approach similar perplexity as centralized.

\section{Partial model training}
\label{app:pvt}
\begin{table}
\centering
\caption{Test perplexity after $10$K communication rounds of training for each class of model and PVT \% of trainable variables.}
\begin{tabular}{cccc}
Model & Trainable \% & \# Parameters & Perplexity \\
\hline
Small LSTM & $100\%$ & $4.7$M & $34.80$ \\
Small Transformer & $100\%$ & $4.1$M &  $38.66$ \\
Small Conformer & $100\%$ & $4.1$M &  $36.80$ \\
\hline
Large LSTM & $100\%$ & $18.8$M & $30.83$ \\
Large LSTM & $40\%$ & $7.5$M & $31.53$ \\
Large LSTM & $20\%$ & $3.8$M & $32.93$ \\
\hline
Large Transformer & $100\%$ & $21.0$M & $29.15$ \\
Large Transformer & $40\%$ & $8.4$M & $30.45$ \\
Large Transformer & $20\%$ & $4.2$M & $32.61$ \\
\hline
Large Conformer & $100\%$ & $20.2$M & $29.06$ \\
Large Conformer & $40\%$ & $8.1$M & $30.06$ \\
Large Conformer & $20\%$ & $4.0$M & $31.51$ \\
\end{tabular}
\label{tab:pvt}
\end{table}

\begin{figure}
\centering
\includegraphics[scale=0.45]{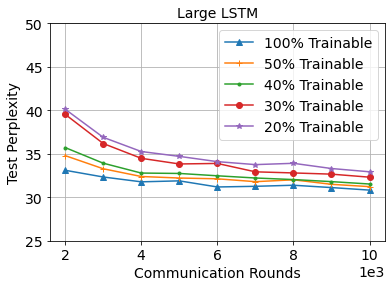}
\includegraphics[scale=0.45]{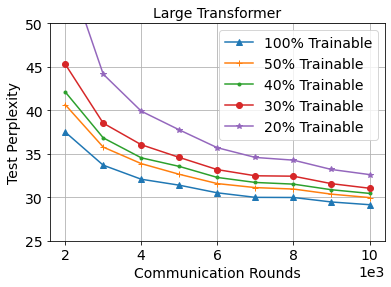}
\includegraphics[scale=0.45]{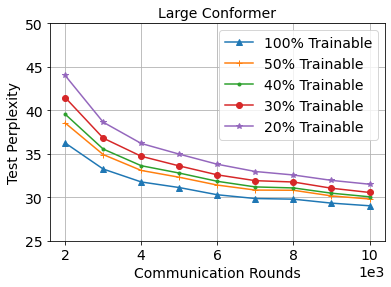}
\caption{Test perplexity over communication rounds for the large models with select percentages of trainable variables denoted by $X\%$ with $100\%$ indicating all trainable variables are trained (i.e. baseline).}
\label{fig:pvt-curve}
\end{figure}

In our experiments with PVT, we vary the percentage of trainable variables from $10\%$ to $90\%$ in increments of $10$.
As before, we search over the hyperparameters in Table~\ref{tab:baseline-hyper} and find them to be mostly consistent with baseline other than client learning rate.
Following \citet{yang2021partial}, we use the per client per round (PCPR) configuration, where the frozen variables vary from round to round and from client to client, as this was shown to achieve the highest accuracy.
Specifically, we only freeze subsets of the multiplicative vectors and matrices of the original model.
This corresponds to the embedding and weights of the LSTM, and for the Transformer and Conformer, the weights of the MLP layer, attention matrices, layer normalization in each block, embedding, and weights for Conformer convolution.
We also note though that although overall the number of trainable variables might average to the desired percentage (e.g. $10\%$), for certain architectures, like LSTM, that don’t have that many \emph{freezable variables} (only one layer’s weight matrix and embedding matrix), the number of trained variables will be much more variable from round to round.
On the other hand, for architectures, like Transformer and Conformer, that have more freezable variables (each blocks’ weight matrices and attention matrices and embeddings), the number of trained is much more consistent between rounds.

We report test set perplexity over communication rounds for the large architectures and varying degrees of PVT in Figure~\ref{fig:pvt-curve} with the number of clients per round set to $800$.
Looking at Table~\ref{tab:pvt}, it is evident that both large models can handle some percentage of partial freezing up until a certain point and that the Large Conformer with only $30\%$ of trainable variables can reach a better perplexity than the Large LSTM with $100\%$ trainable variables by $10$K rounds or so.
However, training for the full $10$K rounds can be a communication bottleneck so PVT would need to be combined with another technique to reduce the number of rounds needed.

\section{Quantization}
\label{app:quant}

In stochastic $k$-level uniform quantization \cite{suresh2017distributed}, values in each layer are converted into one of $k$ evenly distributed values between the layer min and max, stochastically assigned to the closest target value either above or below the real value. The lower the $k$ value, the more the data is being compressed, as the number of bits used to store the value equals $\log_2(k)$. For download quantization, we explore $k$ values corresponding to between $8$ and $28$ bits. For upload quantization, which can be a larger bottleneck in edge devices \citep{mobile-speeds-05-2021}, we explore $k$ values corresponding to between $1$ and $28$ bits. On upload, we also try applying zero-centering during uniform quantization as well as trying the TernGrad \citep{wen2017terngrad} algorithm, which quantizes values in each vector $v$ into only one of three values, $0$ and $\pm\max(|v|)$, corresponding to $\log_2(3)$ ($\sim 1.585$) bits per parameter. While TernGrad is designed to use L infinity clipping ($\ell_\infty$), we experiment with and without this for completeness.

\begin{figure}[t]
\centering
\includegraphics[scale=0.45]{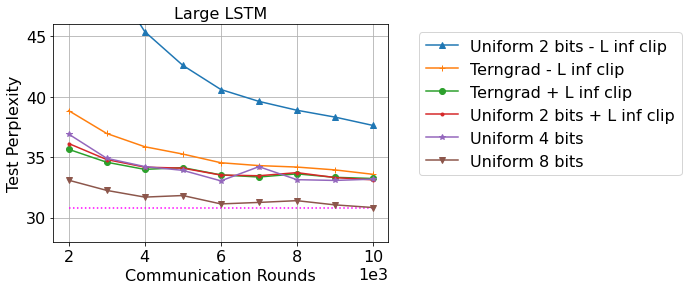}
\includegraphics[scale=0.45]{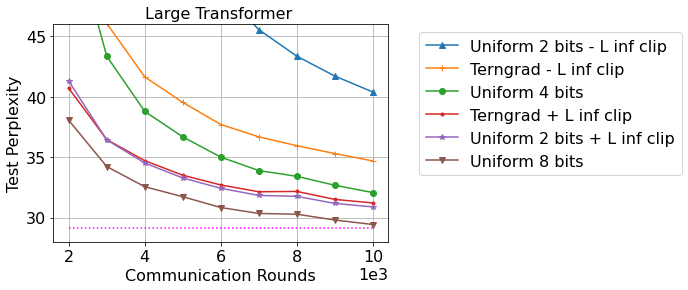}
\includegraphics[scale=0.45]{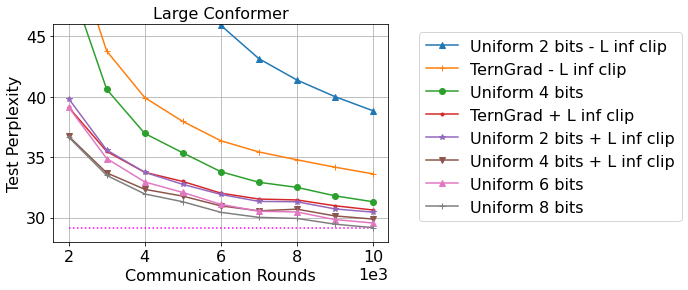}
\caption{Test set perplexity over communication rounds for varying upload quantization levels, with download quantization fixed to $16$ bits. The dotted line shows baseline perplexity achieved after $10$K rounds without any quantization.}
\label{fig:quant_upload_detailed}
\end{figure}

While $\ell_\infty$ clipping did make a significant difference in the TernGrad experiment for Transformers and Conformers, performing much better with it than without, it did not have a large effect on the TernGrad performance in the LSTM in Figure~\ref{fig:quant_upload_detailed}. TernGrad and its counterpart uniform quantization to $\sim1.585$ bits performed the same, as long as $\ell_\infty$ clipping was applied. It is clear from the uniform $2$-bit experiments as well that $\ell_\infty$ clipping is important when quantizing into these lower number of bits; the $2$-bit experiment without clipping performs much worse than the Terngrad without clipping, although enabling clipping allows $2$-bit to perform slightly better than Terngrad's $\log_2(3)$ bits with clipping. Zero-centering did not seem to affect upload behavior much for either model, marginally improving the LSTM and marginally degrading the Transformer.

We explore the patterns of communication cost for each experiment setting in Figure~\ref{fig:quant_comm_costs}. We calculate the approximate download and upload MB for each experiment by multiplying the model's number of parameters by the number of download or upload bits to get total bits transported.

Examining Figure~\ref{fig:quant_comm_costs}, we note the baseline points for each set of experiments as the lowest and rightmost, getting the best perplexity but also highest communication cost. Starting from there, we see trends of no perplexity degradation as we apply conservative quantization to the Large LSTM and Transformer and Conformer settings and move left in the plot. We then reach an elbow in the points for each setting right around where the Terngrad point is, from which point perplexity degrades drastically without much communication cost savings as the points head up in two lines as upload quantization is reduced, with one line corresponding to experiments with download $16$ bits and the other to download $12$ bits. While the Terngrad point for the Large Transformer falls at the outermost point in the "elbow" and therefore gives the best tradeoff for cost versus perplexity, there is one uniform quantization point that does better than the Large LSTM Terngrad, which is download $12$ bits and upload $6$ bits. It makes sense that this does well as we saw that the LSTM was able to use these settings without much regression from the baseline performance, while the Transformer and Conformer could only quantize to $16$ download bits and $8$ upload bits without regressions. 

\section{Transfer learning}
\label{app:transfer}

\begin{table}[ht]
\centering
\caption{Selected hyperparameters for each centrally trained model and dataset.
The values in $[\ ]$ are the possible hyperparameter values searched over.}
\begin{tabular}{ccccc}
Model & Dataset & Clipnorm & Learning Rate \\
& & $[0, 16]$ & $[1e^{-5}, 5e^{-5}, 1e^{-4},$  \\
& & & $5e^{-4}, 1e^{-3}, 5e^{-3}, 1e^{-2}]$ \\
\hline
Large LSTM & Book & $0.0$ & $5e^{-5}$\\
Large LSTM & LM1B & $0.0$ & $5e^{-5}$\\
\hline
Large Transformer & Book & $16.0$ & $5e^{-5}$\\
Large Transformer & LM1B & $16.0$ & $5e^{-5}$\\
\hline
Large Conformer & Book & $0.0$ & $5e^{-5}$\\
Large Conformer & LM1B & $0.0$ & $1e^{-4}$\\
\end{tabular}

\label{tab:central-hyper}
\end{table}

To find the best models pretrained on the Books and LM1B datasets, we train for $30$M steps of synchronous SGD searching over learning rate and clip norm.
Like our other centrally trained models, the batch size is fixed to $16$ and Adam is used with $\beta_1$ at $0.9$, $\beta_2$ at $0.999$, and epsilon at $1e^{-8}$.
See Table~\ref{tab:central-hyper} for the selected hyperparameters.

Next we warmstart each models with the parameters from the best corresponding pretrained centralized model and train using FedAvg for $10$K rounds.
We sweep over clip norm and client learning rate.
See Table~\ref{tab:transfer} for the selected hyperparameters.
Clip norm is omitted in Table~\ref{tab:transfer}, since for all hyperparameter sweeps $16$ was the best value. The Book dataset outperforms the LM1B dataset in all model architectures across LSTM, Transformer, and Conformer. Investigating the difference between the two datasets and their similarities to the Stackoverflow dataset to determine why Books always outperformed LM1B remains an interesting open question.

\begin{table}[h]
\centering
\caption{Test set metrics after $10$K communication rounds of training with $800$ clients per round for each class of model and pretrain dataset. The client learning rate listed is the best performing learning rate found from a hyperparameter sweep. Reported $\Delta$ metrics are the change in quality relative to  Table~\ref{tab:baseline}.}
\begin{tabular}{cccc}
Model & Dataset & \ Client Learning Rate & $\Delta$ Perplexity \\
& & [0.01, 0.1, 0.5, 1.0, 2.0] & \\
\hline
Large LSTM & Book & $0.5$ & $0.76$ \\
Large LSTM & LM1B & $0.5$ & $1.05$ \\
\hline
Large Transformer & Book & $0.1$ & $\mathbf{-0.43}$ \\
Large Transformer & LM1B & $0.1$ & $\mathbf{-0.32}$ \\
\hline
Large Conformer & Book & $0.1$ & $\mathbf{-0.38}$ \\
Large Conformer & LM1B & $0.1$ & $\mathbf{-0.23}$ \\
\end{tabular}
\label{tab:transfer}
\end{table}

\section{Different optimizers}
\label{app:comm-opt}

\begin{table}
\centering
\caption{Test perplexity after $10$K communication rounds of training for each class of model and federated algorithm.}
\begin{tabular}{ccc}
Model & Algorithm & Perplexity \\
\hline
Large LSTM & FedAvg & $30.83$ \\
Large LSTM & MimeLite & $31.00$ \\
Large LSTM & FedProx & $30.76$ \\
\hline
Large Transformer & FedAvg & $29.15$ \\
Large Transformer & MimeLite & $30.39$ \\
Large Transformer & FedProx & $29.04$ \\
\hline
Large Conformer & FedAvg & $29.03$ \\
Large Conformer & MimeLite & $30.41$ \\
Large Conformer & FedProx & $28.93$ \\
\end{tabular}
\label{tab:comm-opt}
\end{table}

In an effort to improve communication efficiency of the larger language models, we examine two communication-efficient federated algorithms: MimeLite and FedProx.
By comparing the speed and point of convergence of these algorithms in number of rounds, we can determine if the overall communication cost of training can be decreased.
As before, we fix the model architectures for each class of model and conduct a basic search over learning hyperparameters using the same common search space as Table~\ref{tab:baseline-hyper} with the addition of the following algorithm specific hyperparameter sweeps.
For MimeLite, we use Adagrad \citep{duchi2011adagrad} for the base optimizer as this setup was shown to perform the best by \citet{karimireddy2020mime} for Stack Overflow.
For the MimeLite Adagrad base optimizer, we sweep over base learning rates of $[0.01, 0.03, 0.1, 0.3, 1.0]$ and epsilons of $[1e^{-1}, 1e^{-3}, 1e^{-5}, 1e^{-7}]$ and fix the server learning rate to $1.0$.
For FedProx, we sweep over $\mu$ values of $[0, 0.1, 0.01, 0.001, 0.0001]$ which controls the weight of the L2 squared norm.

We report test perplexity over $10$K federated training rounds with $800$ clients per round in Figure~\ref{fig:comm-opt} and Table~\ref{tab:comm-opt}.
While FedProx does slightly outperform FedAvg, it does not significantly alter the speed of training in terms of number of communication rounds.
Thus, we chose to continue using FedAvg in the combination experiments for consistency across experiments and more accurate comparisons.

\section{Combination of techniques}
\label{app:combo}

\begin{table}
\centering
\caption{Test perplexity and total communication costs in gigabytes after $10$K communication rounds of training for each class of model and setup. If the number of download bits is unspecified, the standard $32$ bits was used.}
\begin{tabular}{cccc}
Model & Download Cost (GB) & Upload Cost (GB) & Perplexity \\
\hline
Small LSTM & $188$ & $188$ & $34.80$ \\
Small Transformer & $164$ & $164$ & $38.66$ \\
Small Conformer & $162$ & $162$ & $36.80$ \\
\hline
Large LSTM & $752$ & $752$ & $30.83$ \\
Large Transformer & $840$ & $840$ & $29.15$ \\
Large Conformer & $808$ & $808$ & $29.06$ \\
\hline
Efficient Large LSTM (download $32$ bits) & $$752$$ & $75$ & $32.57$ \\
Efficient Large Transformer (download $32$ bits) & $840$ & $84$ & $30.83$ \\
Efficient Large Conformer (download $32$ bits) & $808$ & $81$ & $30.37$ \\
\hline
Efficient Large LSTM (download $16$ bits) & $376$ & $75$ & $32.76$ \\
Efficient Large Transformer (download $16$ bits) & $420$ & $84$ & $32.32$ \\
Efficient Large Conformer (download $16$ bits) & $404$ & $81$ & $31.71$ \\
\end{tabular}
\label{tab:combo}
\end{table}

\begin{figure}
\centering
\includegraphics[scale=0.5]{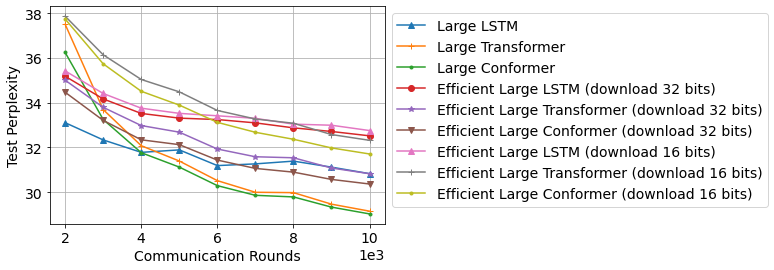}
\caption{Test perplexity over communication rounds for the large models with and without efficient techniques applied.}
\label{fig:combo-curve}
\end{figure}

For the combination experiments, we conducted a joint search over a smaller range of hyperparameters for each technique to keep the total search space reasonable.
For PVT, we restricted the possible percentages to $20\%$, $30\%$, and $40\%$ of trainable variables as those were shown to yield good performance while cutting model size to less than half the original size.
For uniform quantization, we restricted the search of upload to $6$ or $8$ bits and download to $16$ or $32$ bits since the Transformer was shown to be able to handle aggressive upload quantization but required more care on download quantization.
Finally, for transfer learning, we warmstarted after pretraining on the Books corpus.
As in previous experiments, we also search over the common hyperparameter space defined in Table~\ref{tab:baseline-hyper}, where applicable.

Similar to previous experiments, we use $800$ clients per round and train for $10$K rounds with FedAvg.
Figure~\ref{fig:combo-curve} and Table~\ref{tab:combo} contain the results for the large models with and without the efficient techniques applied.
We apply two levels of quantization on download, $16$ and $32$ bits, and observe that the Large LSTM is more amenable to download quantization compared to the Large Transformer and Conformer as the regression between the two levels is much smaller for the LSTM than the Transformer and Conformer.
However, the Transformer and Conformer with $16$ bit download quantization still outperforms all efficient LSTMs though it requires more communication rounds to do so than the efficient Transformer and Conformer with $32$ bits for download.
For the remaining analysis, we focus on the efficient Transformer and Conformer using $32$ bits for download.
It is clear that for the Large Transformer and Conformer, applying efficient techniques yields better quality in earlier communication rounds.
Although there are regressions in the final model quality after $10$K rounds of training, this could be attributed to previously observed issues with increased amounts of labeled data diminishing the value pretraining \citep{rethinkingpretraining2020}.
However, the Efficient Large Transformer and Efficient Large Conformer still reach the same or better final perplexity as the Large LSTM which had no efficient techniques applied.
Furthermore, when considered in terms of actual communication cost, as is done in Figure~\ref{fig:combo-upload}, the efficient models yield much better performance at smaller total communication costs.

\end{document}